\documentclass[11pt,a4paper]{article}
\usepackage{times,latexsym}
\usepackage{url}
\usepackage[T1]{fontenc}

\usepackage{enumitem}

\usepackage{todonotes}
\usepackage{mathtools}
\usepackage{tcolorbox}
\usepackage{amsmath}
\usepackage{amssymb}
\usepackage{array}
\usepackage{multirow}
\usepackage{booktabs}
\usepackage{hyperref}
\usepackage{algorithm}
\usepackage{algpseudocodex}
\usepackage{listings}
\usepackage{float}
\usepackage[nameinlink]{cleveref}
\usepackage{todonotes}
\crefname{section}{\S}{\S\S}
\crefformat{section}{\S#2#1#3}
\usepackage{tabu}
\usepackage{xcolor,colortbl}
\newcommand{\framework}[1]{\textsc{#1}\xspace}

\newcommand{\ourframework}{\framework{Qraft}}

\lstdefinestyle{prompt_code}{
    backgroundcolor=\color{lightgray!50},   
    commentstyle=\color{orange},
    keywordstyle=\color{pink},
    numberstyle=\tiny\color{black},
    stringstyle=\color{orange},
    basicstyle=\ttfamily\footnotesize,
    breakatwhitespace=false,         
    breaklines=true,                 
    captionpos=b,                    
    keepspaces=true,                 
    numbers=left,                    
    numbersep=5pt,                  
    showspaces=false,                
    showstringspaces=false,
    showtabs=false,                  
    tabsize=2
}

\newcommand{\cross}[1][1pt]{\ooalign{%
  \rule[1ex]{1ex}{#1}\cr% Horizontal bar
  \hss\rule{#1}{.7em}\hss\cr}}% Vertical bar

\usepackage{graphicx,calc}
\newlength\myheight
\newlength\mydepth
\settototalheight\myheight{Xygp}
\newcommand*\inlinegraphics[1]{%
  \settototalheight\myheight{Xygp}%
  \settodepth\mydepth{Xygp}%
  \raisebox{-\mydepth-2pt}{\includegraphics[height=\myheight+5pt]{#1}}%
}

\usepackage[acceptedWithA]{tacl2021v1}
\usepackage{xspace,mfirstuc,tabulary}

\newif\iftaclinstructions
\taclinstructionsfalse % AUTHORS: do NOT set this to true
\iftaclinstructions
\renewcommand{\confidential}{}
\renewcommand{\anonsubtext}{(No author info supplied here, for consistency with
TACL-submission anonymization requirements)}
\newcommand{\instr}
\fi

\iftaclpubformat % this "if" is set by the choice of options

\else

\fi

\title{Can LLMs Automate Fact-Checking Article Writing?}

\author{
  \\
  \textbf{Dhruv Sahnan $^1$ \quad David Corney $^2$ \quad Irene Larraz $^3$ \quad
  Giovanni Zagni $^4$} \\
  \textbf{Ruben Miguez $^3$ \quad Zhuohan Xie $^1$ \quad Iryna Gurevych $^{1,5}$ \quad Elizabeth Churchill $^1$} \\
  \textbf{Tanmoy Chakraborty $^6$ \quad Preslav Nakov $^1$}
  \\
  $^1$ MBZUAI, UAE \quad $^2$ Full Fact, UK \quad
  $^3$ Newtral, Spain \quad $^4$ Pagella Politica, Italy \\ 
  $^5$ TU Darmstadt, Germany \quad $^6$ IIT Delhi, India \\
  \texttt{dhruv.sahnan@mbzuai.ac.ae} \quad \texttt{david.corney@fullfact.org} \\ 
  \texttt{irene.larraz@newtral.es} \quad \texttt{ruben.miguez@newtral.es} \\
  \texttt{zhuohan.xie@mbzuai.ac.ae} \quad \texttt{churchill@acm.org\ \ \ \ \ \ \ } \\
  \texttt{\ \ \ \ \ \ \ \ tanchak@iitd.ac.in} \quad \texttt{preslav.nakov@mbzuai.ac.ae}
  \\
}

\date{}

\begin{document}
\maketitle

\begin{abstract}
Automatic fact-checking aims to support professional fact-checkers by offering tools that can help speed up manual fact-checking.
Yet, existing frameworks fail to address the key step of producing output suitable for broader dissemination to the general public:  while human fact-checkers communicate their findings through fact-checking articles, automated systems typically produce little or no justification for their assessments.
Here, we aim to bridge this gap. 
In particular, we argue for the need to extend the typical automatic fact-checking pipeline with \textit{automatic generation of full fact-checking articles}.
We first identify key desiderata for such articles through a series of interviews with experts from leading fact-checking organizations.
We then develop \ourframework{}, an LLM-based agentic framework that mimics the writing workflow of human fact-checkers.
Finally, we assess the practical usefulness of \ourframework{} through human evaluations with professional fact-checkers.
Our evaluation shows that while \ourframework{} outperforms several previously proposed text-generation approaches, it lags considerably behind expert-written articles. We hope that our work will enable further research in this new and important direction.
The code for our implementation is available at \url{https://github.com/mbzuai-nlp/qraft.git}.
  
\end{abstract}

\section{Introduction}

\begin{figure*}[t]
    \centering
    \includegraphics[width=0.9\linewidth]{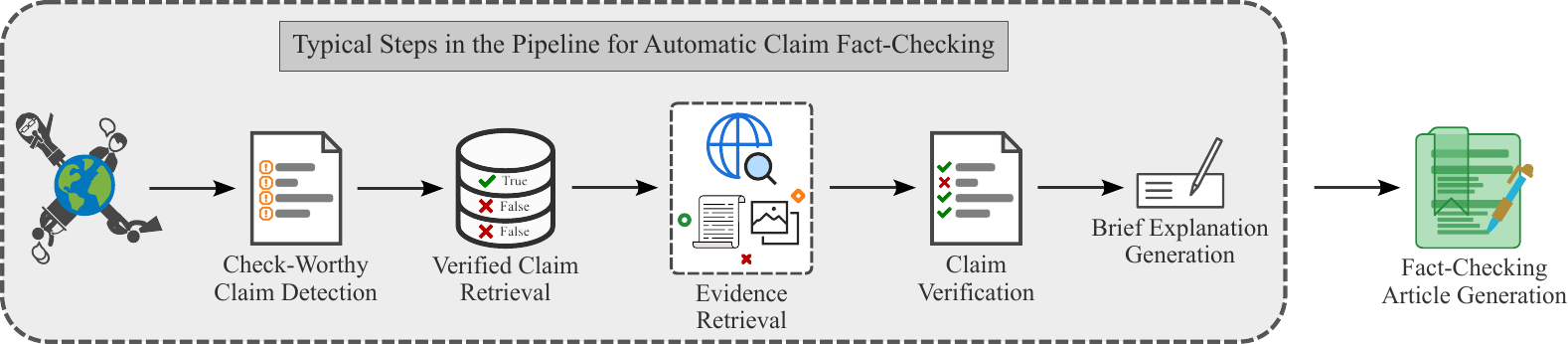}
    \caption{\textbf{Our proposed pipeline for automatic fact-checking.} We extend the typical steps in the automatic claim fact-checking pipeline to include a new task: \emph{fact-checking article generation}.}
    \label{fig:proposed_afc_framework}
\end{figure*}
According to a March 2024 survey~\cite{pewresearch_survey}, nearly 80\% of American adults on major social media platforms regularly encounter news-related content. 
While these platforms allow the rapid spread of \textit{breaking news}, they have also been heavily misused to circulate dubious claims.

In response, the role of fact-checkers has grown to be increasingly vital.
However, manual fact-checking efforts cannot match the scale of global dis/misinformation, prompting a push towards automating the process \cite{vlachos-riedel-2014-fact}.
As shown in \Cref{fig:proposed_afc_framework}, most existing studies frame automatic fact-checking to comprise five main steps: (\emph{i})~check-worthy claim detection, (\emph{ii})~verified claim retrieval, (\emph{iii})~evidence retrieval, (\emph{iv})~claim verification, and (\emph{v})~brief explanation generation~\cite{vlachos-riedel-2014-fact,assist-fc-survey-preslav,CLEF2021:previously:factchecked:claims,guo-etal-2022-survey}. 
Note that some studies skip (\emph{i}), (\emph{ii}) and (\emph{v}).

In practice, fact-checkers do more than just verifying claims; they also publish detailed \textit{fact-checking articles} that guide readers toward a clear understanding of a claim by presenting factual arguments and explaining how they lead to the final verdict~\cite{graves-anatomyofFC-2017}.
Yet existing automatic fact-checking pipelines fail to account for this key time-consuming step of the manual fact-checking workflow. One may argue that \textit{brief explanation generation} serves as the automated counterpart to fact-checking article writing; however, expert requirements for high-quality articles transcend the traditional definition of fact-checking explanations~\cite{warren2025show}. Such explanations are typically concise and aim to justify the factuality decision~\cite{kotonya-toni-2020-explainable}. In contrast, fact-checking articles usually aim not just to provide the evidence leading to the verdict, but also to give the readers the necessary background and contextual information, to explain the origins of the claim, to clarify all likely interpretations of the claim, to address counter-arguments, etc.
We thus argue that the typical  automatic fact-checking pipeline needs to be extended with an extra step, \textit{automatic generation of a fact-checking article}, as shown in \Cref{fig:proposed_afc_framework}.

Moreover, it has repeatedly been highlighted that automatic fact-checking overlooks insights from professional fact-checkers~\cite{assist-fc-survey-preslav,juneja-mitra-2022,das-humancentredfc-2023}.
Thus, we study the task of generating fact-checking articles in close collaboration with experts from several world-leading fact-checking organizations, aiming to understand the key elements of a good fact-checking article.
We further provide a brief analysis of why professional fact-checkers generally do not trust Artificial Intelligence (AI) to fully automate this task. Finally, based on what we have learned from this study, we develop \textbf{\ourframework{}}, a framework using multiple AI agents that collaborate to write and iteratively refine a fact-checking article, designed as the first attempt at a comprehensive solution to cater to the requirements highlighted by the experts.

Our framework leverages large language models (LLMs) as its foundation, as they can generate fluent, long-form text~\citep{yang-etal-2023-doc, xie-etal-2023-next, wang2024generating}. Recent studies have highlighted that agentic workflows can further enhance performance, particularly for tasks involving the generation of documents grounded in a set of facts~\citep{DBLP:journals/corr/abs-2410-02603, shao-etal-2024-assisting,wang2024autopatent}.
Thus, we design \ourframework{} as a multi-agent collaboration that mimics the fact-checking article writing process of human experts. \ourframework{} breaks the writing process down into two main stages. In the first stage, 
\ourframework{} gathers evidence nuggets relevant to the claim, formulates an outline, and then populates it to produce an initial draft. 
In the second stage, \ourframework{} simulates an editorial review that uses conversational question-answering interactions between LLM agents to formulate a list of edits to refine the draft and to ensure professional standards of writing.

We further benchmark \ourframework{} against several state-of-the-art text generation frameworks using automatic evaluation strategies.

In addition, we conduct evaluations with professional fact-checkers focusing on assessing \ourframework{}'s real-world usefulness.
We find that \ourframework{} shows better performance on all automatic metrics; yet, the human evaluations reveal key limitations, indicating the need for constant expert supervision when using LLM-based frameworks for this task.

To sum up, our contributions are as follows:
\begin{itemize}
    \item We introduce the novel task of automatically generating full fact-checking articles, working in close collaboration with experts from leading fact-checking organizations to enrich the task with their insights.
    \item We propose \textbf{\ourframework{}}, a unique framework designed to generate a fact-checking article given (\emph{i})~a claim, (\emph{ii})~its veracity, and (\emph{iii})~a set of evidence documents; we further show that \ourframework{} outperforms several generic text generation approaches.
    \item We conduct expert evaluations to assess the practical usefulness of \textbf{\ourframework{}} and find that it falls short of expert-written articles.
    We also discuss the key limitations 
    as 
    highlighted
    by experts to facilitate future research.
 \end{itemize}

\section{What Expert Fact-Checkers Want from a Fact-Checking Article}
\label{sec:background}

\begin{quote}
    \emph{``
    A fact-checking article for a claim is a form of communication from a fact-checker to the public that provides the necessary context around the claim, argues its veracity, and explains why it may or may not be exactly as claimed.}'' \\\hfill-- An experienced fact-checker
\end{quote}

It is crucial for fact-checkers to maintain public trust in the fact-checking process, which means that they need to be extremely careful when reporting data towards fact-checking a claim.
\newpage

\begin{table}[t]
\centering
\resizebox{1.0\linewidth}{!}{
\begin{tabular}{p{0.3\linewidth}|p{0.8\linewidth}} 
 \toprule
 \textbf{Characteristic} & \textbf{Comments}\\
 \midrule
 Accurately clarifies the claim &
  \textbf{P3} $\xrightarrow{}$``\emph{We must have clarity regarding the veracity of the claim and understand the reasons behind it after reading the article.}''
  \newline
  \textbf{P2} $\xrightarrow{}$``\emph{The article must accurately convey why the evidence contradicts the claim, providing all the necessary context.}''
  \\
  \midrule
 Origin of the claim &
  \textbf{P4} $\xrightarrow{}$``\emph{The article must present the context in which the claim was made, what it meant, and how it affects world events.}''
  \\
 \midrule
  Transparent writing style &
  \textbf{P3} $\xrightarrow{}$``\emph{We must be able to verify everything by consulting the listed sources by ourselves.}''
  \\
  \midrule
  Structure &
  \textbf{P2} $\xrightarrow{}$``\emph{It varies, but generally it begins by introducing the claim, followed by arguments towards its veracity, and ends with a conclusion of our findings.}''
  \\
  \midrule
  Importance of the fact-check &
  \textbf{P1} $\xrightarrow{}$``\emph{Why is the claim worth fact-checking?}''
  \newline
  \textbf{P1} $\xrightarrow{}$``\emph{The article gives context about where the claim was spreading, its harm potential and why it is important to fact-check.}''
  \\
 \bottomrule
\end{tabular}
}
\caption{Summary of the important characteristics of fact-checking articles, according to human experts, along with direct quotes from our interviews with them, explaining what they expected from an article.}
\label{tab:pre-exp-findings-fcs}
\end{table}

To support this, the International Fact-Checking Network has developed a \emph{fact-checkers' code of principles}\footnote{\url{ifcncodeofprinciples.poynter.org}} to streamline the fact-checking process across organizations.
However, fact-checking article writing guidelines still vary between organizations, and individual fact-checkers have their own unique writing styles.
In order to further refine our perception of how fact-checking articles must be composed, we interviewed four experts from world-leading fact-checking organizations.
To maintain the experts' anonymity, we refer to them as P1, P2, P3, and P4.

\subsection{Expert Desiderata for the Articles}
\label{sec:what-do-fact-checkers-want}
We focused our interviews on gathering insights into expert's expectations for a fact-checking article, details that must be present, and preferences about the article's structure and writing style.
Subsequently, we extracted key insights from the responses and categorized them into distinct \emph{characteristics} of a fact-checking article. 

Consistent with \citet{graves-anatomyofFC-2017}, we found that the foremost expectation of a fact-checking article is that it \emph{accurately clarifies} all nuances of the claim, with arguments on how the evidence supports or refutes it, and how they lead to the verdict.
The experts also highlighted that the article must present details on the \emph{origins} of the claim, which includes the necessary background information, details on the context in which it was made, and its implied meaning(s).
Moreover, they emphasized that the article must be written in a \emph{transparent} manner.
P3 elaborated that all arguments made in the article must be supported by publicly available sources, so that readers can verify by themselves that no argument misrepresents information from the original evidence.
These insights echo those by \citet{warren2025show}; however, their work focuses on how fact-checkers construct explanations for a claim's veracity in a fact-checking article.
We further expand these findings to cover the composition of the full fact-checking article.

Adding details toward the \emph{structure} of the fact-checking articles, P2 explained that it varied across organizations, but an article usually began with some background on the claim, including key information regarding the context in which it was spreading.
This is followed by justifications for all likely interpretations of the claim and its proposed veracity assessment, before concluding the article with a summary of the findings.

P1 further revealed that some fact-checking organizations require fact-checkers to specify why the claim was \textit{important} to fact-check.
\Cref{tab:pre-exp-findings-fcs} presents a list of the identified characteristics along with direct quotes from the experts, offering a concise overview of their expectations for a fact-checking article.
The questions we used for the interviews are given in \Cref{appdx:interview_details}.

\subsection{Can AI Do It?}
\label{sec:why_dont_fact_checkers_trust_ai}

We also asked the experts whether they had used AI to assist them in writing fact-checking articles. Most were unaware of any suitable tools for that.
Those who had used general-purpose AI systems said that they had certain limitations, which required human intervention.
Therefore, we collected expert opinions about the limitations they foresaw.
We summarized the key points from their responses and categorized them into anticipated ``issues'' concerning an AI system's capability to perform this task.

\begin{table}[t]
\centering
\resizebox{1.0\linewidth}{!}{
\begin{tabular}{p{0.4\linewidth}| p{0.73\linewidth}} 
 \toprule
 \textbf{Issue} & \textbf{Comments}\\
 \midrule
  Hallucinations &
  \textbf{P1}$\xrightarrow{}$``\emph{LLMs make up arguments and non-existent URLs just to satisfy the veracity of the claim.}''
  \\
 \midrule
  Lack of world-knowledge &
  \textbf{P4}$\xrightarrow{}$``\emph{AI does not have access to world-knowledge like humans do, and thus understanding the relevance of some evidence can be tough.}''
  \\
  \midrule
  Unable to capture context &  
  \textbf{P3}$\xrightarrow{}$``\emph{AI may not be able to put the claim into context because it does not distinguish between what was said and what was implicitly meant.}''
  \\
  \midrule
  Evidence presentation &
  \textbf{P3}$\xrightarrow{}$``\emph{It is hard for AI to have all the context to understand the data and knead it together to argue the claim; instead, it would just give a summary of the evidence as an explanation of the claim's veracity.}''
  \\
 \bottomrule
\end{tabular}
}
\caption{Reasons why experts do not trust an LLM-based system for generating fact-checking articles, together with some direct quotes from our interviews to further detail these issues.}
\label{tab:expected-issues-ai-fcs}
\end{table}

Notably, many of these issues align with well-known challenges associated with LLMs, e.g.,~all experts stated that LLMs were known to \textit{hallucinate} content to make the generated text sound convincing~\cite{hallucinations_survey}. 
P1 mentioned that in his experience, LLMs would make up baseless arguments to satisfy their assessment of the claim's veracity, even though the evidence clearly stated otherwise.
P3 echoed the same concern stating that LLMs cited non-existing URLs as the source for their made up arguments~\cite{chatgpt-hallucinates}. 

Experts also highlighted issues such as LLMs \textit{lacking world knowledge} and likely not being able to \textit{capture the context} in which a claim was made, despite exhaustive evidence being provided~\cite{ko-etal-2024-growover}.
Additionally, P3 mentioned a key shortcoming: he expected AI systems to be unable to construct correct arguments towards the claim's veracity. 
In his experience, AI systems had equated justifying the claim's veracity to a summary of the evidence sources.
However, a fact-checking article must communicate why a claim was assigned a given veracity label, instead of merely summarizing all of the evidence that consulted to arrive with this veracity label.

\Cref{tab:expected-issues-ai-fcs} lists all identified issues, along with direct quotes from fact-checkers providing a brief explanation for their concerns.

\section{Generating Fact-Checking Articles}
\label{sec:methods}

\subsection{Task Definition}
Given a datapoint $X = \{C, V, E_{C,V}\}$, where $C$ represents a claim, $V$ is its veracity, and $E_{C,V}$ is the evidence set consulted for fact-checking the claim, we aim to generate a fact-checking article, $D_{C,V}$, corresponding to $C$ and $V$. 

We assume that the datapoint $X$ is explicitly provided by the fact-checker, and we do not address automatic evidence retrieval or claim verification here.
For evaluating the quality of the generated article, we further have a ground-truth expert-written reference article $D^{gt}_{C,V}$.

\subsection{Methodology}
\label{sec:methodology}
\begin{figure*}[t]
    \centering
    \includegraphics[width=0.88\linewidth]{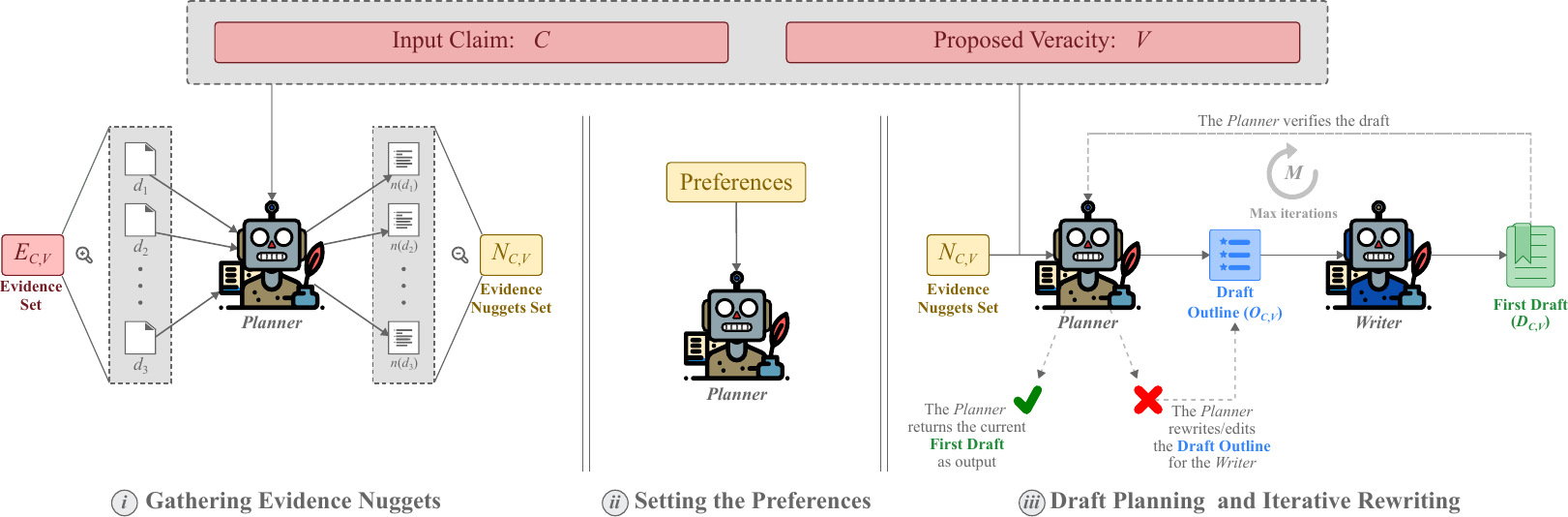}
    \caption{\textbf{Workflow of \ourframework{}: stage (a) --  \textit{planning and compiling the first draft}.} 
    We use two agents, \emph{Planner}     and \emph{Writer}, and we split this stage into three steps: \textit{(i) Gathering Evidence Nuggets}, \textit{(ii) Setting the Preferences}, and \textit{(iii) Draft Planning and Iterative Rewriting}. See \Cref{sec:methodology} for more detail.}
    \label{fig:feedback_diagram_a}
\end{figure*}

\begin{figure}[!t]
    \centering
    \includegraphics[width=0.78\columnwidth]{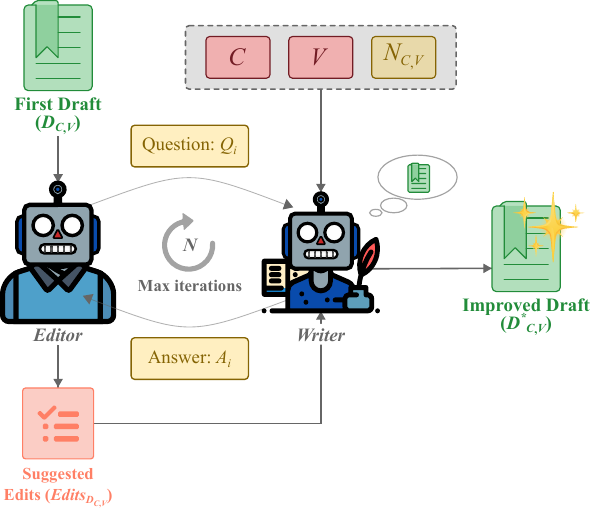}
    \caption{\textbf{Workflow of \ourframework{}: stage (b) -- \textit{simulating an editorial review}.} 
    We simulate conversational question-answering between the \emph{Editor} and the \emph{Writer} in order to generate feedback on how to improve the first draft from \ourframework's stage (a).}
    \label{fig:feedback_diagram_b}
\end{figure}

Experienced writers generally approach any long-form expository writing task by first gathering relevant pieces of evidence (evidence nuggets) from various sources.
These nuggets allow the writers to create an outline that defines the overall structure of their draft and specifies the data to be included in each section~\cite{Rohman_1965}.
This is followed by populating the outline with content, forming the initial draft.
Subsequently, the draft is subjected to proofreading and copy-editing, which are critical steps to refine the writing quality  before publication for public consumption~\cite{Wates_Campbell_2007}.

Here we have a similar approach. We propose \ourframework, an agentic pipeline that replicates the human process of organizing and structuring information for the writing task. 
\ourframework uses conversational interactions between three AI agents as the backbone of the pipeline:
\begin{itemize}
    \item \textit{Planner} $\boldsymbol{\mathcal{P}}$, which plays the role of an \textit{AI assistant}, responsible for gathering evidence nuggets from a given evidence set and planning an outline for the draft.
    \item \textit{Writer} $\boldsymbol{\mathcal{W}}$, which plays the role of a \textit{human fact-checker}, responsible for composing a draft for the fact-checking article.
    \item \textit{Editor} $\boldsymbol{\mathcal{E}}$, which plays the role of a \textit{human expert editor}, responsible for reviewing and helping to refine the drafts composed by $\boldsymbol{\mathcal{W}}$.
\end{itemize}

We illustrate the workflow of \ourframework in \Cref{fig:feedback_diagram_a} and \Cref{fig:feedback_diagram_b}, which show the two main stages: 
\newpage

\noindent \textit{(a) Planning and Compiling the First Draft}, where the article's structure is defined and an initial draft is composed, and \textit{(b) Simulating Editorial Review}, where the draft is iteratively refined using feedback from simulated writer $\xleftrightarrow{}$ editor interactions.
We present pseudocode for the implementation of \ourframework in \Cref{appdx:qraft_pseudocode}.

\subsubsection*{(a) Planning and Compiling the First Draft}
We further break this stage into three broad steps as shown in \Cref{fig:feedback_diagram_a}, which we detail below.
 
\paragraph{\inlinegraphics{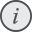} Gathering Evidence Nuggets.}

In this step, the \emph{Planner} $\boldsymbol{\mathcal{P}}$ extracts key evidence relevant to the claim $C$ and its veracity $V$ from each document $d \in E_{C,V}$ and summarizes it into concise bullet points, denoted by $n(d)$. 
This results in the formation of a set of evidence nuggets $N_{C,V}$:

\begin{equation*}
    N_{C,V} = \{n(d)|\ \forall\ d\ \in E_{C,V}\}
\end{equation*}

\paragraph{\inlinegraphics{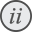} Setting Preferences.}
In order to encourage \ourframework{} to generate articles that align with the \emph{characteristics} of a fact-check article outlined in \Cref{sec:what-do-fact-checkers-want}, we design a list of preferences comprising high-level guidelines about the expected structure and content of the draft.
We provide these preferences to $\boldsymbol{\mathcal{P}}$ in the form of an instruction prompt.

\paragraph{\inlinegraphics{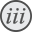} Draft Planning and Writing.}
Here, the \emph{Planner} $\boldsymbol{\mathcal{P}}$ first proposes an outline, $O_{C,V}$, aligning with the preferences specified earlier, using $N_{C,V}$ as the source of evidence nuggets relevant to the claim.
The \emph{Writer} $\boldsymbol{\mathcal{W}}$ then uses $C$, $V$, and $N_{C,V}$ to populate $O_{C,V}$, resulting in the first draft, $D_{C,V}$. 
Note that $\boldsymbol{\mathcal{W}}$ does not access the preferences directly; instead, $\boldsymbol{\mathcal{P}}$ implicitly encodes them into the outline $O_{C,V}$.
This design simplifies the \emph{Writer}'s task, allowing it to focus solely on expanding the outline into an article, without needing to redundantly process the preferences again.

Afterwards, the \emph{Planner} $\boldsymbol{\mathcal{P}}$ verifies whether the compiled draft $D_{C,V}$ aligns with the preferences set in the previous step. Based on this assessment, $\boldsymbol{\mathcal{P}}$ may either approve $D_{C,V}$ for the second stage of \ourframework's workflow or it may self-reflect and refine $O_{C,V}$ to ensure that a draft constructed over a modified outline adheres more closely to our preferences. The interactions in this step are repeated until $\boldsymbol{\mathcal{P}}$ approves the draft $D_{C,V}$ or for a maximum of $M = 5$ iterations.

\subsubsection*{(b) Simulating an Editorial Review}

As illustrated in \Cref{fig:feedback_diagram_b}, in this stage, we simulate conversational interactions between the \textit{Editor} $\boldsymbol{\mathcal{E}}$ and the \textit{Writer} $\boldsymbol{\mathcal{W}}$, to derive feedback for improving the first draft $D_{C,V}$.
$\boldsymbol{\mathcal{E}}$ begins by identifying unclear parts in the first draft $D_{C,V}$ and formulating questions for $\boldsymbol{\mathcal{W}}$ seeking clarification.

This initiates a conversational question-answering based interaction session between the two LLM agents.
As part of each interaction $i$, $\boldsymbol{\mathcal{E}}$ generates a question $Q_{i}$ based on $D_{C,V}$ and the interaction history $\{Q_1, A_1, \dots, Q_{i-1}, A_{i-1}\}$. 
$\boldsymbol{\mathcal{W}}$ then responds with an answer $A_i$ for $Q_i$. The session is conducted for a total of $N = 10$ such interactions, after which $\boldsymbol{\mathcal{E}}$ generates a list of suggested edits $Edits_{D_{C,V}}$ to improve the draft. 
Once the interaction session concludes, $\boldsymbol{\mathcal{W}}$ receives $Edits_{D_{C,V}}$ and applies it on $D_{C,V}$ to produce the improved draft $D^{*}_{C,V}$.
The entire workflow of this stage is repeated on the improved draft $D^{*}_{C,V}$ returned at the end of each cycle. The iterative process continues until the editor determines that no further edits are required, or until a maximum of $K = 5$ improvement cycles have been completed.

For the purpose of our experiments, we use \texttt{GPT-4o-mini} for Planner $\boldsymbol{\mathcal{P}}$ and Writer $\boldsymbol{\mathcal{W}}$.
The Editor $\boldsymbol{\mathcal{E}}$, who is responsible for guiding the Writer to improve its draft, is instantiated using \texttt{GPT-4o}.
This configuration balances efficiency and performance by using a lightweight low-cost model for the resource-intensive task of long-form draft generation, and reserving a more capable LLM for the editorial phase, where high-quality critique is crucial.

\begin{table}[t]
\centering
\resizebox{\linewidth}{!}{
\begin{tabular}{@{}lclc@{}} 
 \toprule
 \multicolumn{1}{l}{\textbf{Dataset}} & \multicolumn{1}{c}{\textbf{\# Examples}} & \multicolumn{1}{l}{\textbf{Veracity Labels}} & \multicolumn{1}{c}{\textbf{\# Labels}}\\
 \midrule 
 \multirow{10}{*}{\it \bf ExClaimCheck} & \multirow{10}{*}{987} & \textit{True} (67), \textit{Mostly True} (125), & \multirow{10}{*}{18}\\ 
& & \textit{Half True} (137), \textit{Barely True} (132), \\
& & \textit{False} (380), \textit{Pants on Fire} (64), \\
& & \textit{Mostly False} (1), \textit{Partly False} (8), \\
& & \textit{Mixture} (1), \textit{Misleading} (49), \\
& & \textit{Mixed} (1), \textit{Missing Context} (10), \\
& & \textit{Faux} (1), \textit{Distorts the Facts} (2), \\
& & \textit{Out of Context} (2), \\
& & \textit{False Attribution} (1), \\
& & \textit{No evidence} (2), \textit{Satire} (4). \\
 \midrule
 \multirow{4}{*}{\it \bf AmbiguousSnopes} & \multirow{4}{*}{240} & \textit{Mostly True} (16), \textit{Mixture} (118), & \multirow{4}{*}{7}\\
& & \textit{Mostly False} (24), \\
& & \textit{Misattributed} (28), \textit{Scam} (14), \\
& & \textit{Correct Attribution} (34), \textit{Legit} (6). \\
 \midrule
 \multirow{12}{*}{\textbf{Total}} & \multirow{12}{*}{\textbf{1,227}} & \textbf{\textit{True} (67), \textit{Mostly True} (141),} & \multirow{12}{*}{\textbf{22}}\\
 & & \textbf{\textit{Half True} (137), \textit{Barely True} (132),} & \\
 & & \textbf{\textit{False} (380), \textit{Pants on Fire} (64),} & \\
 & & \textbf{\textit{Mostly False} (25), \textit{Partly False} (8),} & \\
 & & \textbf{\textit{Mixture} (119), \textit{Misleading} (49),} & \\
 & & \textbf{\textit{Mixed} (1), \textit{Missing Context} (10),} & \\
 & & \textbf{\textit{Misattributed} (28), \textit{Faux} (1),} & \\
 & & \textbf{ \textit{Correct Attribution} (34),} & \\
 & & \textbf{\textit{False Attribution} (1), \textit{Scam} (14),} & \\
 & & \textbf{\textit{Legit} (6), \textit{Out of Context} (2),} & \\
 & & \textbf{\textit{No evidence} (2),Satire (4),
 } & \\
 & & \textbf{\textit{Distorts the Facts} (2).} & \\
 \bottomrule
\end{tabular}
}
\caption{\textbf{Statistics about the \emph{test} dataset we used to evaluate \ourframework{}:} 
number of examples and veracity labels (with frequency of each such label, shown in parentheses).}
\label{tab:dataset-stats-complete}
\end{table}

\section{Experimental Setup}
\subsection{Datasets}
\label{subsec:datasets}

We used two datasets, which we merged into one:

\paragraph{\emph{ExClaimCheck}} \cite{zeng-gao-2024-justilm} contains claims from publicly accessible fact-checking websites (published between November 2008 and March 2021), together with a veracity label, a set of webpages used as evidence, and the full expert-written fact-checking article. We used 987 examples from its test set, for which the fulltext of the evidence articles was available. 

\paragraph{\emph{AmbiguousSnopes}} is a fresh and hard dataset that we curated. It comprises 240 claims, along with their veracity labels, evidence documents, and the expert-written fact-checking article from \emph{Snopes}\footnote{\url{https://www.snopes.com}} (published between October 2022 and August 2024).  We only kept hard claims that could not be judged outright as completely True or completely False.
This was suggested by the experts: that a real-world system must also excel at claims that are challenging even for expert fact-checkers.

\Cref{tab:dataset-stats-complete} gives statistics about the datasets and the inventory of labels they used. Below, we report evaluation results on the union of the two datasets: a total of 1,227 examples using 22 veracity labels.

\subsection{Baselines}
We compare \ourframework{} to several
text-generation approaches.
To the best of our knowledge, there is no prior work on generating entire fact-checking articles.
Therefore, we compare to several prior methods and baselines as described below.

\paragraph{Na\"{i}ve.} 
We collected the top-$k$ semantically similar sentences (to the claim $C$) from each document in the evidence set, then we concatenated them to form a candidate fact-checking ``article.'' 
We set $k = 3$ for the purpose of this experiment.

\paragraph{Summarization.}
Several previous studies modeled the generation of fact-checking \emph{explanations} as a text summarization task~\cite{atanasova-etal-2020-generating-fact,xing-etal-2022-automatic}, and thus wanted to try this as a baseline for generating fact-checking \emph{articles}.
In particular, we fine-tuned \textit{PRIMERA}~\cite{Xiao_Beltagy_Carenini_Cohan_2022} using the evidence webpages as the source and the corresponding fact-checking article as the target (for this, we used additional 5,964 examples from the training split of \emph{ExClaimCheck}).

\begin{table*}[t]
\scriptsize
\centering
\resizebox{\textwidth}{!}{  
    \begin{tabular}{l|ccc|c|c|c|c}
    \toprule
    \multicolumn{1}{l|}{\multirow{2}{*}{\textbf{Method}}} & \multicolumn{3}{c|}{\textbf{ROUGE $\uparrow$}} & \multicolumn{1}{c|}{\multirow{2}{*}{\textbf{BERTScore $\uparrow$}}} & \multicolumn{1}{c|}{\textbf{Entailment}} & \multicolumn{1}{c|}{\multirow{2}{*}{\textbf{FactScore $\uparrow$}}} & \multicolumn{1}{c}{\textbf{Hallucinated}}\\
    & R$_1$ & R$_L$ & R$_{Lsum}$ & & \textbf{Score $\uparrow$} & & \textbf{Citations (\%) $\downarrow$}\\
    \midrule
    Na\"{i}ve Top-$k$ & 0.20 & 0.09 & 0.09 & 0.83 & 0.20 & \textbf{\textit{NA}} & \textbf{\textit{NA}}\\
    PRIMERA & 0.26 & 0.11 & 0.11 & 0.83 & 0.14 & 0.20 & \textbf{\textit{NA}}\\
    JustiLM & 0.07 & 0.05 & 0.05 & 0.80 & 0.23 & 0.55 & \textbf{\textit{NA}}\\
    Vanilla-GPT & 0.29 & 0.13 & 0.16 & 0.79 & \textbf{0.34} & 0.74 & 2.30 \\
    Storm & 0.35 & 0.13 & 0.20 & 0.82 & 0.32 & 0.72 & 1.63\\
    \midrule
    \rowcolor{black!15}
    \multirow{1}{*}{\textbf{\ourframework{}(a)}} & 0.36 & 0.13 & 0.19 & 0.84 & 0.29 & 0.81 & 3.51\\ 
    \rule{0pt}{10pt}\multirow{1}{*}{\textit{\quad-- w/o evidence nuggets}} & 0.32 & 0.13 & 0.18 & 0.83 & 0.28 & 0.80  & 3.62\\
    \multirow{1}{*}{\textit{\quad-- w/o draft verification}} & 0.31 & 0.13 & 0.18 & 0.84 & 0.29 & 0.81 & $\cross[.6pt]$\\
    \multirow{1}{*}{\textit{\quad-- w/o outline}} & 0.31 & 0.12 & 0.18 & 0.84 & 0.27 & 0.79 & 3.33\\
    \multirow{1}{*}{\textit{\quad-- w/o preferences}} & 0.34 & 0.13 & 0.18 & \underline{\textbf{0.85}} & 0.27 & 0.81 & $\cross[.6pt]$\\[2.5pt]
    \rowcolor{black!15}\multirow{1}{*}{\textbf{\ourframework{}(a) + (b)}} & \textbf{0.38} & \underline{\textbf{0.14}} & \textbf{0.21} & \underline{\textbf{0.85}} & 0.30 & \textbf{0.83} & \textbf{1.29}\\
    \rule{0pt}{10pt}\multirow{1}{*}{\textit{\quad-- w/o question-asking}} & 0.36 & \underline{\textbf{0.14}} & 0.20 & \underline{\textbf{0.85}} & 0.29 & 0.81 & 1.94\\
    \bottomrule
    \end{tabular}
    }
    \caption{\textbf{Automatic evaluation.} 
    $\uparrow$ and $\downarrow$ indicate that a higher or a lower score is better, respectively. \textbf{\textit{NA}} denotes that the measure is not applicable for that method, and $\cross[.6pt]$ signals the absence of in-text citations. A \textbf{bold} score represents the best performance for an evaluation measure, while an \underline{\bf underlined bold} is a tie for the best results.}
    \label{tab:untrained_automatic_metrics}
\end{table*}

\paragraph{Justification.}
We used \textit{JustiLM}~\cite{zeng-gao-2024-justilm}, a preexisting method from the literature for justification generation, which uses Atlas~\cite{Izacard_Lewis_Lomeli_Hosseini_Petroni_Schick_Dwivedi-Yu_Joulin_Riedel_Grave_2022} as the base large language model. 
We modified it to generate full fact-checking articles (it uses a mini-fine-tuning on 30 examples randomly sampled from training) to align with the requirements of our task.

\paragraph{LLM-based approaches.} \
We used the following two LLM-based approaches: 
(\emph{i})~\textit{Vanilla-GPT}, where we prompted GPT-4o-mini~\cite{openai2024gpt4technicalreport} to generate a full fact-checking article given the input in a zero-shot setting, and 
(\emph{ii})~\textit{Storm}~\cite{shao-etal-2024-assisting}, which is an agentic long-form text generation pipeline, designed to write Wikipedia-like articles from scratch. 
We retained the workflow of Storm, but we modified it to generate fact-checking articles by customizing the prompts in the underlying components, and restricting the retrieval component to use the evidence documents from our dataset.

\section{Automatic Evaluation}
\label{sec:automatic_eval}
Below, we report the evaluation results for our \ourframework{} framework in comparison to the methods described in the previous section; we also perform ablation studies. 

Our evaluation focuses on assessing whether the generated text aligns with the characteristics of the fact-checking articles outlined in \Cref{sec:what-do-fact-checkers-want} and whether it exhibits traits that exacerbate the experts' distrust in AI, as identified in \Cref{sec:why_dont_fact_checkers_trust_ai}.

\subsection{Evaluation Measures}
\label{automatic_eval_metrics}

We used several automatic evaluation measures, namely \textit{ROUGE}\footnote{We report ROUGE-1, ROUGE-L and ROUGE-Lsum} \cite{Lin_2004} and  
\textit{BERTScore} \cite{Zhang_Kishore_Wu_Weinberger_Artzi_2020} to measure the lexical and the semantic similarity of the generated full fact-checking articles with respect to the references, and \textit{FactScore}\footnote{For FactScore, we restricted the knowledge base to the expert-written articles from our evaluation datasets.}~\cite{Min_Krishna_Lyu_Lewis_Yih_Koh_Iyyer_Zettlemoyer_Hajishirzi_2023} for evaluating their factuality.

We further used the percentage of \textit{Hallucinated Citations}, which calculates the proportion of URLs cited in the text that do not exist in the claim's evidence set, and \textit{Entailment Score}, which measures the consistency and the coverage of the generated fact-checking article $D_{C,V}$ with respect to the ground-truth $D^{gt}_{C,V}$~\citep{zeng-gao-2024-justilm}. The latter measure calculates the mean of SummaC scores~\cite{summac_tacl} over the ordered pairs $(D_{C,V}, D^{gt}_{C,V})$ and  $(D^{gt}_{C,V}, D_{C,V})$.

Moreover, we performed LLM-as-a-judge evaluations~\cite{llm_as_a_judge-2023} on the generated articles, assessing them with respect to
\textit{Relevance},
\textit{Comprehensibility},
\textit{Importance},
and \textit{Evidence presentation}.
For this, we collaborated with the fact-checking experts to design five-point rubrics to score the generated articles on each of these aspects (see \Cref{appdx:llm_as_a_judge} for rubric details). 
Since we evaluated on custom criteria, we selected Prometheus-2~\cite{kim2024prometheus}, an LLM tuned for such assessments, as the judge.

\subsection{Performance Comparison}
\label{sec:results_and_discussion}
\paragraph{Automatic evaluation} 
\begin{table}[t]
\scriptsize
\resizebox{\linewidth}{!}{
    \centering
    \begin{tabular}{@{\ \ }l|c@{\ \ \ }c@{\ \ \ }c@{\ \ }|c@{\ \ \ }c@{\ \ \ }c@{ }}
    \toprule
    \multirow{2}{*}{\textbf{Method}} & \multicolumn{3}{c|}{\textbf{ExClaimCheck}} & \multicolumn{3}{c}{\textbf{AmbiguousSnopes}} \\
    & ES $\uparrow$ & FS $\uparrow$ & HC (\%) $\downarrow$ & ES $\uparrow$ & FS $\uparrow$ & HC(\%) $\downarrow$ \\
    \midrule
    Na\"{i}ve Top-$k$ & 0.21 & \textbf{\textit{NA}} & \textbf{\textit{NA}} & 0.17 & \textbf{\textit{NA}} & \textbf{\textit{NA}} \\
    PRIMERA & 0.12 & 0.19 & \textbf{\textit{NA}} & 0.23 & 0.22 & \textbf{\textit{NA}} \\
    JustiLM & 0.24 & 0.56 & \textbf{\textit{NA}} & 0.19 & 0.51 & \textbf{\textit{NA}} \\
    Vanilla-GPT & \textbf{0.34} & 0.74 & 1.89 & \textbf{0.33} & 0.72 & 3.93 \\
    Storm & \textbf{0.34} & 0.73 & 1.58 & 0.26 & 0.71 & 1.83 \\
    \midrule
    \textbf{\ourframework{}(a)} & 0.30 & 0.81 & 3.86 & 0.27 & \textbf{0.81} & 2.09 \\
    \textbf{\ourframework{}(a) + (b)} & 0.31 & \textbf{0.85} & \textbf{1.25} & 0.29 & \textbf{0.81} & \textbf{1.46} \\
    \bottomrule
    \end{tabular}
    }
    \caption{\textbf{Automatic evaluation on individual datasets.} \textbf{\textit{NA}} denotes that the measure is not applicable for that method. A \textbf{bold} score represents the best performance for an evaluation measure.}
    \label{tab:dataset_spec_results}
\end{table}
\Cref{tab:untrained_automatic_metrics} presents a performance comparison of various frameworks across multiple evaluation measures. We can see that LLM-based approaches consistently achieve better performance.

\ourframework{} outperforms the other LLM-based approaches, surpassing them on 6 out of 7 evaluation measures.
Notably, \ourframework{} achieves the highest FactScore, demonstrating 11 points of improvement over the next-best method, while also having the least amount of hallucinated citations. 
\textit{Storm} emerges as the best baseline with scores within 3 points of the highest on each measure, except for FactScore.
\textit{Vanilla-GPT} follows closely behind, maintaining competitive scores; however, it exhibits a relatively higher number of hallucinated citations, and underperforms both on BERTScore and ROUGE.
On deeper analysis, we also find that it cites only about 30\% of the available evidence sources on average, whereas \ourframework{} cites more than 90\% of all evidence sources in the article it generates.
Moreover, from \Cref{tab:dataset_spec_results}, we can see that models maintain similar performance across both datasets.
\textit{Vanilla-GPT} produces a relatively much higher proportion of hallucinated citations on \textit{AmbiguousSnopes} at 3.93\%---compared to 1.89\% on \textit{ExClaimCheck} -- while \textit{Storm} shows degradation in the Entailment Score on \textit{AmbiguousSnopes}.
\ourframework{} largely remains stable across both datasets, with slight degradations in performance on \textit{AmbiguousSnopes}.

\paragraph{LLM-as-a-judge evaluation} 
\begin{table}[t]
\scriptsize
\resizebox{\linewidth}{!}{
    \centering
    \begin{tabular}{l|c|c|c|c}
    \toprule
    \textbf{Method} & \textbf{Rel} & \textbf{Com} & \textbf{Imp} & \textbf{Evi}\\
    \midrule
    Na\"{i}ve Top-$k$ & 2.31 & 1.77 & 1.96 & 1.60 \\
    PRIMERA & 1.13 & 1.09 & 1.21 & 1.15 \\
    JustiLM & 1.21 & 1.14 & 1.07 & 1.05 \\
    Vanilla-GPT & 4.31 & 4.33 & 4.21 & 4.11 \\
    Storm & 4.65 & 4.11 & 4.13 & 4.32 \\
    \midrule
    \multirow{1}{*}{\ourframework{}(a)} & 4.52 & 4.15 & 4.49 & 4.25 \\
    \multirow{1}{*}{\textbf{\ourframework{}(a) + (b)}} & \textbf{4.70} & \textbf{4.77} & \textbf{4.54} & \textbf{4.56} \\
    \bottomrule
    \end{tabular}
    }
    \caption{\textbf{LLM-as-a-judge evaluation.} We report scores on a five-point scale for four aspects: \textit{Relevance} (Rel), \textit{Comprehensibility} (Com), \textit{Importance} (Imp), and \textit{Evidence presentation} (Evi).
    }
    \label{tab:llm_as_a_judge}
\end{table}

\Cref{tab:llm_as_a_judge} presents results for LLM-as-a-judge evaluations on the four aspects we discussed above. 
These experiments reinforce the trend that LLM-based frameworks outperform other baselines on this task. 

\textit{Storm} outperforms \textit{Vanilla-GPT}, benefiting from its multi-perspective question-asking approach. 
This lets \textit{Storm} conduct a more thorough analysis, leading to a clearer explanation of the claim and detailed, but sometimes overly elaborate, evidence presentation.

Regardless of these strengths, \ourframework{} still emerges as the preferred method, demonstrating the best scores across all four aspects.
The editorial review simulated in the second stage of our framework enables it to focus on clarifying the claim by arguing its truthfulness with evidence without presenting unnecessary details. 

\subsection{Ablation Studies}
\label{sec:qraft_ablations}

Considering that \ourframework{} decomposes the process of generating fact-checking articles into multiple steps (\Cref{sec:methods}), it is natural to question the need for each step.
\Cref{tab:untrained_automatic_metrics,tab:dataset_spec_results,tab:cost_efficiency,tab:llm_as_a_judge} include some ablation experiments, which we discuss below.\footnote{In \Cref{tab:cost_efficiency,tab:dataset_spec_results}, \textbf{ES} denotes Entailment Score; \textbf{FS} denotes FactScore; and \textbf{HC} denotes Hallucinated Citations.}

\paragraph{\ourframework{}'s stages.} We observed that dropping stage (b) from our \ourframework{} framework led to degradation across all automatic  measures, including aspects of the articles assessed through LLM-as-a-judge evaluations. 
Most notably, there is a 2.7 times increase in the proportion of hallucinated citations and 0.62 point absolute drop in \textit{Comprehensibility}, which is likely due to the absence of an \emph{Editor} to help the \emph{Writer} clarify and refine its arguments.
However, it is important to highlight that our \ourframework{} framework (a) is still on par with the best baseline on most aspects, and even outperforms it on FactScore by 0.09.

\paragraph{Steps within each stage.} We assessed four variations of \ourframework{}(a):
(\emph{i})~\emph{w/o evidence nuggets}, where we skip the compression of evidence into concise nuggets,
(\emph{ii})~\emph{w/o draft verification}, where the \emph{Planner} does not verify the draft compiled by the \emph{Writer},
(\emph{iii})~\textit{w/o outline}, where the \emph{Writer} generates the draft without an outline,
and (\emph{iv})~\textit{w/o preferences}, where no preferences are provided.

We observee that each of these variations showed slight degradations across all evaluation measures.
Interestingly, dropping preferences or draft verification resulted in the complete absence of any in-text citations using URLs.
On closer examination, we found that the evidence was instead cited using webpage titles without URLs, leaving no way for the reader to verify whether the information was correctly presented. 
Moreover, with respect to \ourframework{}(b), we found that using it \emph{w/o question-asking} yields to slight degradation across all evaluation measures with a notable increase in the hallucinated citations. 

\begin{table}[t]
\scriptsize
\resizebox{\linewidth}{!}{
    \centering
    \begin{tabular}{@{\ \ }l|c|c|c|c@{\ \ }}
    \toprule
    \textbf{Method} & \textbf{Avg. Cost} & \textbf{ES $\uparrow$} & \textbf{FS $\uparrow$} & \textbf{HC (\%) $\downarrow$} \\
    \midrule
    \rowcolor{black!15}
    \textbf{\ourframework{}(a)} & & & & \\
    \textit{\quad-- draft verification} & & & & \\
    \textit{\quad\ \ \ $M = 0$} & \$0.0036 & 0.29 & 0.81 & $\cross[.6pt]$ \\
    \textit{\quad\ \ \ $M = 1$} & \$0.0050 & \textbf{0.30} & 0.80 & 4.17 \\
    \textit{\quad\ \ \ $M = 3$} & \$0.0064 & 0.28 & 0.81 & 7.03 \\
    \textit{\quad\ \ \ $M = 5$} & \$0.0096 & 0.29 & 0.81 & 3.51 \\
    \textit{\quad\ \ \ $M = 7$} & \$0.0125 & 0.29 & 0.81 & 4.04\\
    \rowcolor{black!15}
    \textbf{\ourframework{}(a) + (b)} & & & & \\
    \textit{\quad-- question-asking} & & & & \\
    \textit{\quad\ \ \ $N = 0$} & \$0.049 & 0.29 & 0.81 & 1.94 \\
    \textit{\quad\ \ \ $N = 3$} & \$0.054 & 0.27 & 0.82 & 1.93 \\
    \textit{\quad\ \ \ $N = 7$} & \$0.058 & 0.29 & \textbf{0.83} & 1.97 \\
    \textit{\quad\ \ \ $N = 10$} & \$0.063 & \textbf{0.30} & \textbf{0.83} & 1.29 \\
    \textit{\quad\ \ \ $N = 12$} & \$0.067 & \textbf{0.30} & \textbf{0.83} & 1.40 \\
    \textit{\quad\ \ \ $N = 15$} & \$0.070 & 0.29 & \textbf{0.83} & \textbf{1.23} \\
    \textit{\quad-- editorial review} & & & & \\
    \textit{\quad\ \ \ $K = 1$} & \$0.012 & 0.27 & 0.81 & 1.72 \\
    \textit{\quad\ \ \ $K = 3$} & \$0.037 & 0.28 & 0.82 & 1.52  \\
    \textit{\quad\ \ \ $K = 5$} & \$0.063 & \textbf{0.30} & \textbf{0.83} & 1.29 \\
    \textit{\quad\ \ \ $K = 7$} & \$0.076 & \textbf{0.30} & \textbf{0.83} & 1.36\\
    \bottomrule
    \end{tabular}
    }
    \caption{\textbf{Cost and efficiency analysis.} $\cross[.6pt]$ signals the absence of in-text citations. A \textbf{bold} score represents the best performance for an evaluation measure.}
    \label{tab:cost_efficiency}
\end{table}

\paragraph{Cost and efficiency.} 
Since \ourframework{} comprises multiple cycles of LLM invocations with long-form articles, we assessed the average cost of the framework against the quality of the generated articles across three variables: the number of \textit{draft verification} cycles $M$, the number of \textit{question-asking} interactions $N$, and the number of \textit{editorial review} cycles $K$.\footnote{We fixed $M=5$ and $K=5$ while varying $N$, and we set $M=5$ and $N=10$ while varying $K$.}

We can see that
for
\ourframework{}(a), increasing $M$ raises the cost per generated article gradually from $\$0.003$ to $\$0.012$, offering a 1.2--2$\times$ reduction in the proportion of hallucinated citations, while the performance on other metrics remains stable.
Introducing \ourframework{}(b) incurs higher costs -- up to $\$0.076$ -- but leads to substantial performance gains, especially, a much lower percentage of hallucinated citations (a best of 1.23\%).

We note that varying the number of question-asking interactions ($N$) is relatively inexpensive, and provides improvements in the form of higher factual accuracy and a reduced amount of hallucinated citations. 
Increasing the number of editorial review cycles ($K$) is costlier, but yields consistent performance gains across all metrics.
We choose $M=5$, $N=10$, and $K=5$ as the default configuration for \ourframework{} because this setup produces high-quality articles while maintaining cost-efficiency.

\section{Expert Evaluation}
\label{sec:expert_eval}
While automatic evaluations indicate that \ourframework{} outperforms existing methods, they are insufficient to assess the real-world usefulness of the generated fact-checking articles for expert fact-checkers. 
Several studies have demonstrated ROUGE and BERTScore to be less reliable in single-reference settings~\cite{sheng-etal-2024-reference}. 
In particular, due to their reliance on lexical or semantic similarity-based text overlap, their ability to capture the diversity of expression in text is limited.
Perfectly acceptable fact-checking articles can be written for the same claim in multiple different ways. However, because there is only one human-written reference per claim in the dataset, these metrics could unfairly treat generated articles as of lower quality, solely due to stylistic differences.
FactScore also exhibits several limitations, such as overestimating factuality in certain cases~\cite{chiang-lee-2024-merging}, or assuming all claims in the generated text are verifiable~\cite{song-etal-2024-veriscore}.
These limitations, combined with the fact that FactScore is specifically optimized to assess the factuality of generated biographies using Wikipedia as evidence, render the metric insufficient to evaluate the performance of LLM-based frameworks on our task.
Moreover, LLM-based evaluation is not infallible: LLMs are known to be biased towards LLM-generated text assigning them higher scores~\cite{justic_or_prejudice_llm_bias}.

Furthermore, LLM-as-a-judge may not reflect expert preferences due to the specialized nature of the task and our need for specifically tailored evaluation criteria~\cite{limitations_llm_as_a_judge_iui_2025,huang2025empiricalstudyllmasajudgellm}.
The usefulness of the generated text extends beyond these automatic evaluation strategies, requiring expert judgment to assess whether the fact-check articles align with professional standards.

\begin{table}[t]
\scriptsize
\resizebox{\linewidth}{!}{
    \centering
    \begin{tabular}{l|c|c|c|c|c}
    \toprule
    \textbf{Method} & \textbf{Rel} & \textbf{Com} & \textbf{Imp} & \textbf{Evi} & \textbf{Pub}\\
    \midrule
    Vanilla-GPT & 4.41 & 3.91 & 2.83 & 3.62 & 3.16 \\
    Storm & 3.75 & 3.50 & 3.41 & 3.29 & 2.00\\
    \multirow{1}{*}{\ourframework{}(a) + (b)} & 4.47 & 4.16 & \textbf{3.75} & 3.45 & 3.25\\
    \midrule
    \multirow{1}{*}{\textbf{Expert}} & \textbf{5.00} & \textbf{4.66} & 3.25 & \textbf{4.79} & \textbf{4.83}\\
    \bottomrule
    \end{tabular}
    }
    \caption{\textbf{Expert evaluation.} We report average scores on a five-point scale for five aspects: \textit{Relevance} (Rel), \textit{Comprehensibility} (Com), \textit{Importance} (Imp), \textit{Evidence presentation} (Evi), and \textit{Publishability} (Pub). 
    }
    \label{tab:expert_evals}
\end{table}

We conducted human evaluations with the help of our expert fact-checkers.
We designed a questionnaire to rate the fact-checking articles using a 5-point Likert scale on the aspects listed in \Cref{automatic_eval_metrics}, along with \textit{Publishability}, which served as a proxy to measure the quality and the usefulness of the generated articles (see \Cref{appdx:user_study_questionnaire} for questionnaire details).
We manually collected 12 claims related to climate change, which were 
fact-checked by \emph{different} fact-checking organizations, and generated automatic fact-checking articles for these claims using \textit{Vanilla-GPT}, \textit{Storm}, and \ourframework{}.
We then presented each of our experts with a claim and asked them to blindly rate the corresponding fact-checking articles: the three automatic ones and the original article.
Finally, we asked the expert fact-checkers to rank these articles by relative \textit{Publishability}.

\begin{figure}[t]
    \centering
    \includegraphics[width=\linewidth]{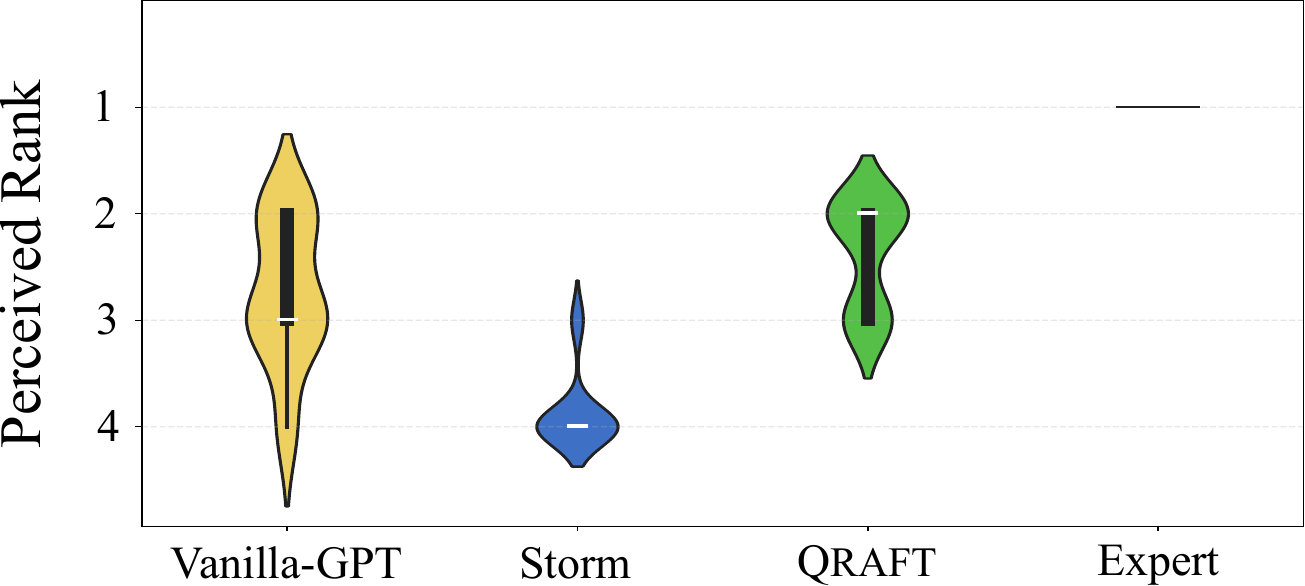}
    \caption{Distribution of the rankings received by the four fact-check articles based on relative publishability as perceived by the experts.}
    \label{fig:expert_evals_ranking}
\end{figure}

\Cref{tab:expert_evals} presents the scores for the five evaluation aspects, while \Cref{fig:expert_evals_ranking} illustrates the distribution of the rankings for the four articles. We can see that all LLM-based frameworks achieve considerably lower scores than the expert-written articles, which are consistently recognized as the best.
All three generative frameworks are rated within the 3-4 point range, indicating expert uncertainty or considerable room for improvement in quality across most aspects.
In terms of rankings, expert-written articles were clearly found to be of the highest quality, while \textit{Storm}'s generations were ranked the lowest.

This is further reinforced by the \emph{Publishability} scores, where the experts indicated that they would rather write an article from scratch than use \textit{Storm}'s version. 
\ourframework{} was ranked second by most experts, while \textit{Vanilla-GPT} received inconsistent rankings, showing a median rank of third.
This is again reflected in the \emph{Publishability} scores, where the experts expressed uncertainty regarding the potential usefulness of these articles indicating the need for considerable manual editorial effort to raise them up to professional writing standards.

We further asked the experts for open-ended qualitative feedback explaining their scores:

\paragraph{Articles generated by LLM frameworks are too elaborative.}
The experts noted that while \textit{Storm} and \ourframework{} thoroughly discussed the evidence, they often included extraneous details that did not help explain why the claim might or might not be true. These details made the articles harder to read and the arguments less comprehensible. This concern relates to all the issues relayed by experts in the initial interviews: hallucinations, lack of world knowledge, inability to capture context, and poor presentation of evidence.
In addition, when multiple evidence sources offered similar information, these articles tended to be overly repetitive.
Expert-written articles generally consolidate such evidence into a single concise argument, citing all sources to strengthen their case.
\textit{Storm} exhibits this limitation the most, due to its overly elaborate articles, while \ourframework{} is less concerning.

\paragraph{\ourframework{} presents factual information, but makes errors in using it to clarify the claim.}
The experts highlighted that \ourframework{} presented data that was relevant to the claim; however, it failed to construct arguments using this data to explain the claim's veracity.

For instance, in an article, \ourframework{} reported the amount of CO$_2$ emissions from fossil fuel burning, but did not mention how this related to a significant percentage of atmospheric CO$_2$ being anthropogenic, and not natural as claimed.

Moreover, the articles occasionally presented information from certain evidence sources as facts, while, in reality, this might be only true within a specific context, and does not necessarily constitute \textit{the whole truth}.

This indicates that the articles lack a thorough and accurate analysis, failing to provide all the necessary context around a claim to explain its veracity.

\paragraph{Organizations exhibit considerable diversity in their guidelines for article composition.}
In \Cref{sec:what-do-fact-checkers-want}, we mentioned that the article writing guidelines varied between fact-checking organizations,
as different fact-checking organizations have distinct writing guidelines for their articles, 
and this point was further reinforced by our evaluation.
We asked the experts whether the articles explicitly specified why the claim was ``important'' to fact-check, and we observed low scores for all four fact-checking articles.
The experts clarified that, while even the expert-written articles did not always explain the importance of the fact-check, these would still be considered acceptable by some organizations, although their own organization might have written it differently.
In addition, the experts also voiced their disagreement on the choice of evidence, noting that their organizations would have preferred to use official documents as sources, while others would be more flexible and could use third-party news reportings about the same.

\paragraph{Experts do not trust fully AI-generated articles.}
Fact-checking articles are written in a way that they transparently justify the verdict of a claim by guiding readers through the investigative process~\cite{graves-anatomyofFC-2017}.
Consequently, experts carefully analyze any fact-checking article they encounter and verify for themselves, using the cited evidence, whether all arguments are presented accurately.
They stressed that an AI-generated article requires an even higher level of scrutiny and cannot be trusted as-is, emphasizing that a complete draft generated by \ourframework{} does not save them much time, but offers ideas on possible structures for their own fact-checking article.

Yet, the experts highlighted the potential of an AI framework that generates full fact-checking articles under human supervision. 
In such a framework, fact-checkers could specify their preferences and constantly guide the underlying models at the intermediate steps to align with their requirements and preferences.
This would enable them to quickly prototype initial drafts for their fact-checking article, which they could easily edit and polish for actual publication.

\section{Related Work}
\paragraph{Automatic fact-checking.}
Previous work has focused on automating fact-checking, and not so much on assisting expert fact-checkers. Yet, they decomposed the fact-checking process into useful subtasks, which could help experts~\cite{vlachos-riedel-2014-fact,kotonya-toni-2020-explainable,assist-fc-survey-preslav,guo-etal-2022-survey}: claim check-worthiness detection~\cite{hassan-checkworthy-2015,wright-augenstein-2020-claim,konstantinovskiy-claim-detection-2021,clef2023}, detecting fact-checked claims~\cite{shaar-etal-2020-known}, evidence retrieval~\cite{zhou-etal-2019-gear,10.1007/978-3-030-45442-5_45,zou-etal-2023-decker}, claim verification~\cite{wang-2017-liar,augenstein-etal-2019-multifc,DBLP:journals/corr/abs-2411-00784}, and explanation generation~\cite{popat-etal-2018-declare,atanasova-etal-2020-generating-fact,zeng-gao-2024-justilm}.

However, these tasks overlook the important step of generating fact-checking articles, which is a crucial step of the manual fact-checking workflow~\cite{graves-anatomyofFC-2017}.
It is a time-consuming process, and recent work has advocated for the development of NLP tools to assist experts in this task~\cite{liu-cscw-factcheckers-needs-2024}.
\citet{warren2025show} recently presented insights into how fact-checkers compose explanations for a claim's veracity in fact-checking articles, highlighting features consistent with those that we outline in this work (see \Cref{sec:background}).
Yet, the emphasis of their work is on improving the alignment of automatic fact-checking explanations with the needs of professional fact-checking.
Explanations traditionally serve as a rationale of the underlying model's decision-making process~\cite{kotonya-toni-2020-explainable}, whereas experts write fact-checking articles to communicate their findings and clear the air regarding a claim by guiding readers through the exact investigation process~\cite{graves-anatomyofFC-2017}.
Thus, here we have argued for the need to introduce the \emph{generation of fact-checking articles} as an additional task.

\paragraph{Long-form text generation.}
Long-form text generation is challenging, as it requires ensuring that the generated text remains on topic, follows a natural narrative arc, and preserves the intended meaning. To address these issues, previous studies have broken down the process into stages inspired by human writing processes, including: planning, drafting, rewriting, and editing~\citep{yao2019plan, yang2022re3, hu2022planet, yang-etal-2023-doc, liang2024integrating}.
Moreover, expository writing requires the text to be grounded on external evidence~\cite{weaver1991expository}, which demands a thorough sense-making process over the evidence, along with an ability to collate information into a cohesive narrative~\cite{shen2023summarizationdesigningaisupport}.
While some methods have been proposed~\cite{balepur2023expositorytextgenerationimitate}, recent research has highlighted the effectiveness of agentic frameworks, which emphasize the pre-writing stage~\cite{shao-etal-2024-assisting,wang2024autopatent,Wang_Guo_autosurvey,liu-chang-2025-writing}.
Fact-checking article writing is one such task, and our approach uses LLM agents to structure it into the above four stages of long-form writing.
This ensures that \ourframework{} maintains topical consistency, and clarifies the claim with strong support from the evidence.

\section{Conclusion and Future Work}
Previous automatic fact-checking research has focused on tasks that can potentially assist expert fact-checkers to perform claim verification more efficiently.
However, fact-checkers also perform another largely overlooked, yet essential task: communicating their findings regarding the claim through detailed fact-checking articles.
In this paper, we argued that the typical fact-checking pipeline must be extended to include a new task: \emph{the automatic generation of fact-checking articles}.
We defined this task in close collaboration with experts from leading fact-checking organizations, deepening our understanding with insights into what constitutes a good fact-checking article.
Based on these insights, we proposed \ourframework{}, a multi-agent collaboration framework that mimics the fact-checking article writing process of human experts.
Moreover, through comprehensive evaluation, using automated evaluation measures, LLM as a judge, as well as expert judgments and qualitative analysis, we showed that \ourframework{} outperforms several preexisting approaches from the literature.

As human evaluation revealed that \ourframework{} still falls short compared to expert-written articles, particularly in terms of practical usefulness, we aim to compile a more detailed set of requirements for such articles. Moreover, as existing automatic evaluation measures are insufficient to capture the real-world utility of the generated articles, we aim to develop a more robust suite of evaluation measures for this task. We also aim to benchmark stronger, more recent LLMs for this purpose, especially such tuned for complex reasoning---commonly known as \emph{large reasoning models}.

In our experiments, we further observed that \ourframework{} failed to construct arguments using the evidence to explain the claim's veracity in some cases.
To alleviate this gap, in future work, we aim to introduce an intermediate reasoning step, which would focus on deeper sense-making of the claim and the broader narrative using context provided by the evidence.
Finally, we plan to explore the potential for human--AI cooperation~\cite{subhabrata_haico}, which would enable for higher-quality fact-checking articles thanks to human oversight and iterative feedback.

\section*{Limitations}

Our work introduced a new task into the typical automatic fact-checking pipeline, namely \textit{the automatic generation of full fact-checking articles}.
We also proposed \ourframework{}, a framework for the end-to-end generation of such articles given a claim and the supporting evidence.
Here, we acknowledge some limitations of our work, which remain open research questions for future work.

First, we relied on the inherent capabilities of the underlying LLMs in our framework to preserve relevant context when extracting concise evidence nuggets from source evidence documents.
However, this strategy may not ensure the retention of all context---particularly for complex claims that often depend on indirectly related or contextually distant information.
We have left a thorough analysis of this limitation for future work.

Second, \ourframework{} consists of three LLM agents, which collaborate to generate and iteratively refine full fact-checking articles. 
In our experiments, we exclusively used OpenAI's \texttt{GPT-4o} and \texttt{GPT-4o-mini} as the underlying LLMs.
We have left a systematic performance comparison of alternative model combinations as agents in such a multi-agent framework to future research.

Finally, we designed \ourframework{} with a focus on \textit{assisting professional fact-checkers} in the time-consuming and labor-intensive fact-checking article writing process, instead of generating drafts intended for public dissemination as-is without human oversight.
We expect the users of our framework to be professional fact-checkers, and thus, we only consulted fact-checking experts during the human evaluation phase of our work.
However, an in-depth investigation of how non-experts perceive the generated fact-checking articles is also important and would be interesting to study in future work.

\section*{Acknowledgment}
The authors would like to acknowledge the Multi-Institutional Faculty Interdisciplinary Research Project (MFIRP) between IIT Delhi and MBZUAI and Anusandhan National Research Foundation
(DST/INT/USA/NSF-DST/Tanmoy/P-2/2024) for financial support. T.C. further acknowledges the support of the Rajiv Khemani
Young Faculty Chair Professorship in Artificial Intelligence.

\bibliography{rebibed_custom,tacl2021}
\bibliographystyle{acl_natbib}

\newpage 

\appendix

\section{Details about the Interviews}
\label{appdx:interview_details}
In \Cref{sec:background}, we presented the key findings from our interviews with experts from leading fact-checking organizations.
Here, we present the structure of our interviews along with the questions we asked in \Cref{tab:prelim_interview_questions}.
We kept the questions open-ended in order to keep the process of gathering insights unbiased. 

\begin{table}[t]
\centering
\resizebox{0.9\linewidth}{!}{
\begin{tabular}{p{0.15\linewidth}|p{0.95\linewidth}} 
 \toprule
 \multicolumn{1}{c|}{\textbf{Section}} & \multicolumn{1}{c}{\textbf{Content}}\\
 \midrule 
 \multicolumn{1}{c|}{\textbf{I.}} & \multicolumn{1}{c}{\textbf{General Question Round}} \\
 & \textit{Q.} As a fact-checker, what do you expect from a fact-check article? \\
 & \textit{Q.} Can you comment on any challenges for AI that you may foresee in the process of writing a fact-check article? \\
 \\
 \multicolumn{1}{c|}{\textbf{II.}} & \multicolumn{1}{c}{\textbf{Sample Data Presentation}} \\
 & \textit{Claim:} ``In December 2023, conspiracy theorist <person\_1>'s Twitter/X account, @<username\_p1>, was reinstated by <person\_2> after having been banned for abusive behavior since 2018.''\\
 & \textit{Veracity:} True \\
 & \textit{Evidence:} \{Website 1, Website 2, \dots\}\\
 \\
 \multicolumn{1}{c|}{\textbf{III.}} & \multicolumn{1}{c}{\textbf{Focused Question Round 1}} \\
 & \textit{Q.} What do you expect from an article written for the given data? \\
 & \textit{Q.} What do you think an AI-based system might miss if it generated a fact-check article from the given data? \\
 \\
 \multicolumn{1}{c|}{\textbf{IV.}} & \multicolumn{1}{c}{\textbf{Presenting Generated Article}} \\
 & \multicolumn{1}{c}{\textit{Article:} -- Redacted --} \\
 \\
 \multicolumn{1}{c|}{\textbf{V.}} & \multicolumn{1}{c}{\textbf{Focused Question Round 2}} \\
 & \textit{Q.} What are your opinions on the article based on your expectations? Do you think your expectations were met? \\
 & \textit{Q.} What do you think of the utility of this article for a fact-checker? \\
 \\
 \multicolumn{1}{c|}{\textbf{VI.}} & \multicolumn{1}{c}{\textbf{Concluding Question Round}} \\
 & \textit{Q.} Do you have any other points or any comments on something we might have missed? \\
 \midrule
 \bottomrule
\end{tabular}
}
\caption{\textbf{Structure of our interview along with the questions asked in each round.} Each interview took approximately one hour and was held one-on-one with a fact-checking expert. The article in Section IV was generated using few-shot prompting on GPT-4o-mini.}
\label{tab:prelim_interview_questions}
\end{table}

\section{Details about the QRAFT Implementation}
\label{appdx:qraft_pseudocode}
In \Cref{sec:methods}, we introduced \ourframework{}, our framework designed to rely on conversational question-asking between multiple LLM agents to draft fact-checking articles.
Here, we give some specifics on \ourframework{}'s workflow through a pseudocode in \Cref{algo:qraft_pseudocode}.

\begin{algorithm}[t]
    \scriptsize
    \caption{Pseudocode for \ourframework{}}
    \begin{algorithmic}
        \Require $X = \{C, V, E_{C,V}\}$\\
        \textbf{where} $E_{C,V} = \{d\ |\ d\text{ is an evidence document}\}$\\
        \\
        $\boldsymbol{\mathcal{P}}$: the Planner agent\\
        $\boldsymbol{\mathcal{W}}$: the Writer agent\\
        $\boldsymbol{\mathcal{E}}$: the Editor agent \\
        ${Prefs}$: Guidelines for the draft's structure\\
        \\
        $M$: Max iterations for draft planning and writing\\
        $N$: Max iterations for Question-Asking interactions between $\boldsymbol{\mathcal{W}}$ and $\boldsymbol{\mathcal{E}}$
        \Procedure{\ourframework{} (a)}{}
        \LComment{\inlinegraphics{icons_1_32.png} Evidence nugget from the input by $\boldsymbol{\mathcal{P}}$}
        \State Initialize $N_{C,V} \gets \{\}$
        \ForAll{$d\ \in E_{C,V}$}
            \State $n(d) \gets \boldsymbol{\mathcal{P}}$.gatherEvidenceNuggets$(d, C, V)$
            \State $N_{C,V}$.add$(d:n(d))$
        \EndFor
        \\
        \LComment{\inlinegraphics{icons_2_32.png} Telling $\boldsymbol{\mathcal{P}}$ the preferences for the draft's structure}
        \State $\boldsymbol{\mathcal{P}}$.setPreferences$({Prefs})$
        \\
        \LComment{\inlinegraphics{icons_3_32.png} Planning and writing the first draft by collaboration between $\boldsymbol{\mathcal{P}}$ and $\boldsymbol{\mathcal{W}}$}
        \State Initialize $i \gets 0$
        \State Initialize $O_{C,V} \gets$ `` ''
        \State Initialize $D_{C,V} \gets$ `` ''
        \While{$i < M$}
            \State $O_{C,V} \gets \boldsymbol{\mathcal{P}}$.proposeOutline$(N_{C,V}, C, V, O_{C,V}, D_{C,V})$
            \State $D_{C,V} \gets \boldsymbol{\mathcal{W}}$.writeDraft$(O_{C,V}, C, V, N_{C,V})$
            \If{$\boldsymbol{\mathcal{P}}$.approveDraft$(D_{C,V})$}
            \State break
            \EndIf
            \State $i \gets i+1$
        \EndWhile
        \State \Return $D_{C,V}$
        \EndProcedure
        \\
        \Procedure{\ourframework{} (b)}{}
        \State Initialize $history \gets \{\}$
        \State Initialize $i \gets 0$
        \State $\boldsymbol{\mathcal{E}}$.reviewDraft$(D_{C,V})$
        \While{$i < N$}
        \LComment{Question-Asking based interactions between $\boldsymbol{\mathcal{E}}$ and $\boldsymbol{\mathcal{W}}$}
            \State $Q_i \gets \boldsymbol{\mathcal{E}}$.makeQuestion$(history)$
            \State $A_i \gets \boldsymbol{\mathcal{W}}$.answer$(Q_i, D_{C,V})$
            \State $history$.add$(\{Q_i,A_i\})$
            \State $i \gets i+1$
        \EndWhile
        \State $Edits_{D_{C,V}} \gets \boldsymbol{\mathcal{E}}$.suggestEdits$(history)$
        \State $D^*_{C,V} \gets \boldsymbol{\mathcal{W}}$.improveDraft$(D_{C,V}, Edits_{D_{C,V}},C,V,N_{C,V})$
        \State \Return $D^{*}_{C,V}$
        \EndProcedure
    \end{algorithmic}
    \label{algo:qraft_pseudocode}
\end{algorithm}

\section{Details about the LLM-as-a-Judge Evaluation}
\label{appdx:llm_as_a_judge}
In \Cref{sec:automatic_eval}, we have already explained that we performed LLM-as-a-judge evaluations using Prometheus-2 as the evaluator LLM, and that we collaborated with fact-checking experts to design five-point rubrics to score the generated articles on several aspects.
Here, we present \Cref{tab:llm_as_a_judge_rubrics}, which shows the particular rubrics we used in our evaluation on the set of four specific aspects we addressed: \textit{Relevance}, \textit{Comprehensibility}, \textit{Importance}, and \textit{Evidence Presentation}.

\begin{table*}[t]
\centering
\resizebox{0.9\linewidth}{!}{
\begin{tabular}{p{0.1\linewidth}|p{\linewidth}} 
 \toprule
 \multicolumn{1}{c|}{\textbf{Aspect}} & \multicolumn{1}{c}{\textbf{Question \& Rubrics}}\\
 \midrule 
 \multicolumn{1}{c|}{\textbf{Relevance}} & \textbf{Q. Is the content of the article relevant to the claim and its proposed veracity?} \\
 & Score 1: The content is irrelevant to the claim and/or its proposed veracity. \\
 & Score 2: Most of the content is inconsistent with the proposed veracity of the claim. \\
 & Score 3: Some of the content is consistent with the proposed veracity of the claim, but some is not. \\
 & Score 4: Most of the content is consistent with the proposed veracity of the claim. \\
 & Score 5: The content in the article is consistent with the proposed veracity of the claim. \\
 \midrule
 \multicolumn{1}{c|}{\textbf{Comprehensibility}} & \textbf{Q. Is the article easy to understand and follow for readers without background knowledge on the claim?} \\
 & Score 1: Clearly not, the reader would definitely need additional background knowledge on the claim to understand the article. \\
 & Score 2: Probably not, the claim is hard to understand without background knowledge, which is not present in the article. \\
 & Score 3: Unsure, readers might need some additional background knowledge to understand the article. \\
 & Score 4: Mostly yes, the article contains most of the necessary context about the claim, but some is missing. \\
 & Score 5: Definitely, the article contains all the necessary context about the claim. \\
 \midrule
 \multicolumn{1}{c|}{\textbf{Importance}} & \textbf{Q. Does the article explain why the claim is being fact-checked?} \\
 & Score 1: The article puts no effort into explaining why the claim is being fact-checked. \\
 & Score 2: Mostly not, it requires some effort while reading the article to guess why the claim is being fact-checked. \\
 & Score 3: The reader could infer why the claim is being fact-checked, but it is not explicitly stated. \\
 & Score 4: The article gives some justification for why the claim is being fact-checked, but it could do better. \\
 & Score 5: The article clearly explains why the claim is being fact-checked. \\
 \midrule
 \multicolumn{1}{c|}{\textbf{Evidence Presentation}} & \textbf{Q. Does the article construct arguments using evidence to explain the claim's veracity?} \\
 & Score 1: The article does not discuss the evidence at all. \\
 & Score 2: The article does not create arguments, mostly just summarizes the evidence. \\
 & Score 3: The article mostly creates arguments, but some evidence is summarized with no focus on how it helps explain the claim's veracity. \\
 & Score 4: The article constructs arguments using evidence to explain the claim's veracity, but there are some gaps in logic. \\
 & Score 5: The article constructs accurate arguments using the evidence to explain the claim's veracity. \\
 \bottomrule
\end{tabular}
}
\caption{\textbf{The questions and rubrics used to judge the generated articles in our LLM-as-a-judge evaluations.} We used Prometheus-2 (7B) as the evaluator LLM.}
\label{tab:llm_as_a_judge_rubrics}
\end{table*}

\newpage 
\section{Details about the Expert Evaluation}
\label{appdx:user_study_questionnaire}
In \Cref{sec:expert_eval}, we presented the expert evaluations that we conducted to judge the practical usefulness of the AI-generated articles, along with the expert-written fact-checking article. 
As we explained there, we evaluated these articles across 5 aspects: \textit{Relevance}, \textit{Comprehensibility}, \textit{Importance}, \textit{Evidence Presentation}, and \textit{Publishability}.
Here, in \Cref{tab:expert_evals_questionnaire}, we present the actual questionnaire we used to rate the articles qualitatively and quantitatively.

\begin{table*}[t]
\centering
\resizebox{0.9\linewidth}{!}{
\begin{tabular}{p{0.1\linewidth}|p{\linewidth}} 
 \toprule
 \multicolumn{1}{c|}{\textbf{Aspect}} & \multicolumn{1}{c}{\textbf{Question \& Rubrics}} \\
 \midrule 
 \multicolumn{1}{c|}{\textbf{Relevance}} & \textbf{Q 1. Does the article clearly state the claim that is being fact-checked?} \\
 & Score 1: Definitely not, the claim is not mentioned at all. \\
 & Score 2: Maybe not, the claim is not articulated clearly, and it is hard to tell what is being fact-checked. \\
 & Score 3: Unsure, the claim is not stated clearly. \\
 & Score 4: Mostly yes, the claim is mentioned, but it is hard to find (e.g., maybe because it is not stated clearly in the introductory part of the article). \\
 & Score 5: Definitely, the claim is clearly stated. \\
 & \textbf{Q 2. Does the article clearly state the claim's veracity?} \\
 & Score 1: Definitely not, no mention of the claim’s veracity in the article at all, or the article states a wrong veracity label \\
 & Score 2: Not really, the veracity of the claim is not explicitly stated, but one might be able to guess it with some effort by reading the entire article carefully. \\
 & Score 3: Maybe, the veracity of the claim can be inferred from the contents of the full article, but it is not clearly stated. \\
 & Score 4: Somewhat, the article only states the veracity of the claim in the introduction and/or in the conclusion, but does not discuss it otherwise. \\
 & Score 5: Definitely yes, the article clearly states the veracity of the claim. \\
 & \textbf{Q 3. Do the contents of the article support the proposed veracity assessment of the claim?} \\
 & Score 1: Definitely not, the presented content is irrelevant to the claim and/or its proposed veracity. \\
 & Score 2: Mostly not, most of the content is inconsistent with the proposed veracity of the claim. \\
 & Score 3: Unsure, some of the content is consistent with the proposed veracity of the claim, and some is not. \\
 & Score 4: Mostly yes, most of the content is consistent with the proposed veracity of the claim. \\
 & Score 5: Definitely yes, the content in the article is consistent with the proposed veracity of the claim. \\
 \midrule
 \multicolumn{1}{c|}{\textbf{Comprehensibility}} & \textbf{Q. Do you think the article is easy to understand for readers without background knowledge about the claim?} \\
 & Score 1: Clearly not, the reader would definitely need additional background knowledge on the claim to understand the article. \\
 & Score 2: Probably not, the claim is hard to understand without background knowledge, which is not present in the article. \\
 & Score 3: Unsure, readers might need some additional background knowledge to understand the article. \\
 & Score 4: Mostly yes, the article contains most of the necessary context about the claim, but some is missing. \\
 & Score 5: Definitely, the article contains all the necessary context about the claim. \\
 \midrule
 \multicolumn{1}{c|}{\textbf{Importance}} & \textbf{Q. Does the article explain why the claim is being fact-checked?} \\
 & Score 1: Not at all, the article puts no effort to justify why the claim is being fact-checked. \\
 & Score 2: Mostly not, it requires some effort while reading the article to guess why the claim is being fact-checked. \\
 & Score 3: Unsure, one could infer from the article why the claim needs to be fact-checked, but this is not clearly articulated. \\
 & Score 4: Mostly yes, the article gives some justification about why the claim is being fact-checked, but it could do better. \\
 & Score 5: Definitely yes, the article gives enough justification about why the claim is being fact-checked. \\
 \midrule
 \multicolumn{1}{c|}{\textbf{Evidence Presentation}} & \textbf{Q. How does the article discuss the evidence?} \\
 & Score 1: Does not discuss the evidence at all. \\
 & Score 2: Mostly just summarizes the contents of the evidence sources. \\
 & Score 3: Mostly constructs arguments, but contents of some evidence sources are just summarized with no focus on how they help the claim's veracity assessment. \\
 & Score 4: Constructs arguments, but there are some gaps in their logic. \\
 & Score 5: Constructs accurate arguments towards the claim’s veracity based on the contents of each evidence source. \\
 \midrule
 \multicolumn{1}{c|}{\textbf{Publishability}} & \textbf{Q. Could this article be published as-is or does it need extra work?} \\
 & Score 1: The article is of no use. \\
 & Score 2: The article gives me some ideas about the structure, but I would rather write one from scratch. \\
 & Score 3: The article could serve as a first draft and could save time for me compared to writing a review-ready article from scratch. \\
 & Score 4: The article could be published after some edits. \\
 & Score 5: The article could be published as is. \\
 \bottomrule
\end{tabular}
}
\caption{\textbf{The questionnaire used to judge the four -- three AI-generated and one expert-written -- articles through expert evaluations.} We presented each expert with a claim and asked them to rate the four corresponding fact-checking articles using this questionnaire. For \textit{Relevance}, we calculated the the mean score across the three questions.}
\label{tab:expert_evals_questionnaire}
\end{table*}

\end{document}